%% file: main.tex
\pdfoutput=1

\documentclass[11pt]{article}

\usepackage{styles/acl}

\usepackage{times}
\usepackage{latexsym}
\usepackage[T1]{fontenc}
\usepackage[utf8]{inputenc}
\usepackage{microtype}
\usepackage{inconsolata}
\usepackage{amsmath}
\usepackage{amssymb}
\usepackage{graphicx}
\usepackage{xcolor}
\usepackage{algorithm}
\usepackage{algpseudocode}
\usepackage{booktabs}
\usepackage{multirow}

\usepackage{caption}
\usepackage{subcaption}

\input{sections/9_macros}

\title{\rt: Geometric Priming via Orthogonal Rotation \\ to Improve Language Model Reasoning}

\author{
 \textbf{Aditya Sharma\textsuperscript{1,2,3}},
 \textbf{Christopher J. Pal\textsuperscript{1,2,4},
 \textbf{Amal Zouaq\textsuperscript{1,2,3}}}
\\
\\
 \textsuperscript{1}Polytechnique Montréal,
 \textsuperscript{2}Mila - Quebec AI Institute,
 \textsuperscript{3}LAMA-WeST Lab,
 \textsuperscript{4}Canada CIFAR AI Chair
\\
 \small{
   \textbf{Correspondence:} \href{mailto:aditya.sharma@mila.quebec}{aditya.sharma@mila.quebec}
 }
}

\begin{document}
\maketitle

\input{sections/0_abstract}
\input{sections/1_introduction}

\input{sections/2_related_work}
\input{sections/3_preliminaries}
\input{sections/4_method}

\input{sections/5_experiments}

\input{sections/6_results}
\input{sections/7_conclusion}
\input{sections/8_limitations}

\section*{Acknowledgements}
\label{sec:acknowledgements}
We thank NSERC, Samsung, IVADO, and the Canada First Research Excellence Fund for supporting this work and CIFAR for their support under the Canada CIFAR AI Chair program.

\bibliography{references}

\input{sections/10_appendix}

\end{document}

%% file: sections/9_macros.tex
\newcommand{\reffig}[1]{Figure~\ref{#1}}
\newcommand{\reftbl}[1]{Table~\ref{#1}}
\newcommand{\refsec}[1]{Section~\ref{#1}}

\newcommand{\rt}{Rotate2Think}
\newcommand{\mv}{MATH-Vision (testmini)}

\newcommand{\ein}{e_{\mathrm{in}}}
\newcommand{\ethink}{e_{\mathrm{th}}}
\newcommand{\ethr}{\hat{e}_{\mathrm{th}}}

\DeclareMathOperator*{\argmin}{argmin}
\newcommand{\conicity}{\mathrm{Conicity}}


%% file: sections/0_abstract.tex
\begin{abstract}
Reasoning models achieve strong performance on challenging tasks by generating explicit intermediate reasoning traces before producing a final answer. Yet the internal structure of representation space when reasoning remains poorly understood: how do a model's hidden representations differ during thinking versus the embeddings of the input prompt, and can this structure be exploited to elicit stronger reasoning at inference time?
We show that both input embeddings and thinking embeddings (mean-pooled last-layer hidden states over the prompt and reasoning trace, respectively) exhibit extremely high conicity, with all vectors clustering tightly around a single mean direction. Crucially, these mean input and thinking directions are non-collinear, with thinking embeddings occupying a geometrically distinct region of embedding space across many different models and benchmark tasks. This observation motivates casting the input-to-thinking transition as a rotation problem admitting a closed-form solution via orthogonal Procrustes analysis. We propose Rotate2Think, a training-free method that estimates this rotation from a small set of correctly solved examples and injects the resulting synthetic thinking vector between thinking delimiters at inference time, providing a geometric primer at the onset of the reasoning trace.Evaluated across multiple benchmarks and model families, Rotate2Think improves accuracy in 30 of 32 model–benchmark configurations across mathematics, science, and code tasks, and generalizes zero-shot to multimodal reasoning on MATH-Vision.
\end{abstract}

%% file: sections/1_introduction.tex
\section{Introduction}
\label{sec:introduction}

The advent of chain-of-thought prompting~\citep{wei2022chain} and large-scale reinforcement learning-based post-training \citep{grpo,rlhf} has given rise to a new class of \emph{reasoning models}: language models that produce explicit intermediate reasoning traces, delimited by special thinking tokens, before generating a final answer~\citep{openai2024o1,guo2025deepseek,yang2025qwen3}. These models achieve strong performance on challenging benchmarks spanning mathematical olympiad problems, graduate-level science questions, and programming tasks precisely because they can allocate additional test-time compute to the problem at hand \citep{snell2024scaling}. Yet despite this empirical success, remarkably little is understood about what geometrically distinguishes a model's hidden representations during thinking from those during direct generation.
If the onset of a reasoning trace shapes the trajectory that follows, then steering the model into a better initial representational state could improve reasoning quality without spending additional test-time compute.

This motivates three main research questions about the geometry of reasoning: 
\textbf{RQ1.} How does the geometry of a model's hidden representations evolve from the embeddings of the input prompt to those of the reasoning trace? 
\textbf{RQ2.} If thinking corresponds to a specific geometric structure in representation space, can we prime the \emph{start} of a reasoning trace, providing both reasoning and base models a stronger initial position from which they then reason fully?
\textbf{RQ3.} If the input-to-thinking geometry is a property of the model rather than of the input distribution, does a rotation fit on text reasoning transfer zero-shot to other modalities, such as visual mathematical reasoning?

We approach these questions empirically by extracting and comparing two types of mean-pooled last-layer embeddings for each problem in a reasoning benchmark: the \emph{input embedding} $\ein(q)$, obtained from a single pass over the user query without any generation, and the \emph{thinking embedding} $\ethink(q)$, the mean-pooled hidden state over the \texttt{<think>}\ldots\texttt{</think>} span of a model-generated reasoning trace. This allows us to directly characterize how representational geometry evolves from the prompt to the reasoning trace.
We discover a striking regularity in both embedding sets: they exhibit extremely high \emph{conicity}~\citep{chandrahas2018geometry}, with all vectors clustering tightly around a single mean direction and average cosine similarity to that mean approaching $1.0$. This holds consistently across diverse model families and benchmark domains spanning mathematics, science, and code. 
Crucially, with average cosine similarity of $0.66$ across models and benchmarks tested, the mean direction of the input space ($\mu_{\mathrm{in}}$) and that of the thinking space ($\mu_{\mathrm{th}}$) are consistently misaligned relative to within-space alignment confirming that thinking embeddings occupy a geometrically distinct subspace, even as each space individually is nearly one-dimensional.
Together, these two observations, high within-space conicity and cross-space misalignment, reduce the input-to-thinking mapping to a rotation problem admitting a closed-form solution via orthogonal Procrustes analysis~\citep{schonemann1966generalized}.
We show this rotation can be estimated from a small set of correctly solved examples drawn from held-out datasets, requiring no examples from the target distribution. Further, the resulting transformation reconstructs thinking embeddings with high fidelity across all models and benchmarks tested (cosine similarity $>0.96$, mean square error $<0.05$), validating both the geometric hypothesis and the cross-domain generality of the learned rotation.


We call this approach \textbf{\rt}.
At inference time, we apply the fitted rotation to the mean-pooled input embedding of a new query to produce a synthetic thinking vector $\ethr(q)$, which is injected into the model's context as a single token position between the \texttt{<think>} and \texttt{</think>} delimiters via the model's \texttt{inputs\_embeds} interface. This geometrically primes the onset of the reasoning trace, providing the model a stronger initial representational position from which it can reason. \rt{} is entirely training-free: it requires only a small collection of correctly solved examples from held-out datasets to fit the rotation, which amounts to a single $D \times D$ matrix computed in seconds and generalizes to unseen benchmarks without any target-domain examples.
We make the following contributions:
\begin{enumerate}
\item \textbf{A geometric characterization of reasoning representations.}
We show that input and thinking embeddings both form nearly one-dimensional cones with non-aligned axes across diverse model families and benchmarks, and that a single rotation maps between them with high fidelity (cosine similarity $>0.96$, mean square error $<0.05$ across models and benchmarks tested).
\item \textbf{\rt, a training-free inference-time method.}
We fit an orthogonal Procrustes rotation from correctly solved examples on held-out datasets and inject the resulting synthetic thinking vector between thinking delimiters, geometrically priming the onset of the reasoning trace without any target-domain data or model modification.

\item \textbf{Cross-domain generalization.} A single rotation, fit on held-out datasets, improves accuracy in 30 of 32 model–benchmark configurations across mathematics, science, and code, spanning four model families from 4B to 31B parameters, with no target-domain examples or per-benchmark refitting.

\item \textbf{Zero-shot cross-modal transfer.} Fit purely on text reasoning, the same rotation transfers zero-shot to visual mathematical reasoning on \mv. On Gemma-4 E4B, Base+R2T (38.1\%) exceeds full reasoning-mode accuracy (34.9\%) on a modality the rotation never observed, suggesting the input-to-thinking geometry is a property of the model, not the input distribution.

\end{enumerate}

%% file: sections/2_related_work.tex
\section{Related Work}
\label{sec:related_work}

\noindent \textbf{Test-Time Reasoning and Scaling.}
Chain-of-thought prompting~\citep{wei2022chain} established that intermediate reasoning steps improve LLM performance, inspiring RL-trained reasoning models, o1~\citep{openai2024o1}, DeepSeek-R1~\citep{guo2025deepseek}, Qwen3~\citep{yang2025qwen3}, that generate extended thinking traces. Subsequent work has largely sought to \emph{reshape} these traces: \citet{snell2024scaling} scale compute up via multi-response search and process verifiers; \citet{muennighoff2025s1} manipulate trace length through budget forcing; \citet{aghajohari2025markovian} restructure the reasoning environment itself to achieve linear compute scaling. 
All require either additional compute at inference or gradient updates to the target model. 
\rt{} is orthogonal to these directions: operating on frozen models with no fine-tuning or additional inference cost, it geometrically primes the \emph{start} of the reasoning trace with a single synthetic embedding. As these approaches manipulate the trace itself, \rt{} could naturally complement them by providing a better-positioned initial state; we leave such combinations for future work.

\noindent \textbf{Geometry of Transformer Representations.}
\rt{} rests on the empirical finding that both input and thinking embeddings exhibit extreme conicity; we use the Alignment-to-Mean and conicity metrics of \citet{chandrahas2018geometry}.
\citet{ethayarajh2019contextual} first demonstrated that all layers of BERT, ELMo, and GPT-2 produce anisotropic cone-like representations; \rt{}'s observed conicity of $\approx 1.0$ (see \reffig{fig:conicity_results}) in last-layer means of reasoning models is the extreme end of this spectrum.
\citet{gaorepresentation} identify an underlying mechanism: auto-regressive LMs with weight tying collapse output embeddings into a narrow cone during training.
\citet{razzhigaev2024shape} show that decoder anisotropy peaks in upper layers and varies systematically with architecture and training phase, consistent with the high conicity \rt{} measures at the final layer.
Prior work has proposed differentiable isotropy measures, finding that anisotropy can improve task performance~\citep{rudman2024stable}, while \citet{tsukagoshi2025redundancy} demonstrate that the intrinsic dimensionality of text embeddings is low enough that 25\% of dimensions suffice for most tasks, corroborating the effective rank-one geometry that justifies fitting Procrustes from only a handful of solved examples.
\citet{karkada2026symmetry} show that translation symmetries in language co-occurrence statistics analytically predict manifold structure in model representations, providing a theoretical mechanism for why a systematic input-to-thinking rotation arises from training data structure.
The Platonic Representation Hypothesis~\citep{huh2024platonic} further suggests that representational geometry converges universally with model scale, supporting the generalizability of the learned rotation across problem domains and model families.
Finally, \citet{sun2019rotate} model knowledge-graph relations as rotations in complex vector space, establishing rotation as a natural and expressive transformation between semantic spaces; \rt{} instantiates this intuition in the real-valued mapping from input to thinking geometry in language models.

\noindent \textbf{Representational Analysis and Steering.}
\citet{park2024linear} formalize the linear representation hypothesis, showing that high-level semantic properties are encoded as linear directions in the representation space, providing some theoretical basis for treating the thinking embedding as a meaningful direction toward which the input embedding can be rotated.
Representation Engineering \citep{zou2023representation} extracts population-level concept directions and adds them to intermediate activations to monitor and control model behavior.
\citet{turner2023steering} similarly construct contrastive steering vectors injected at chosen hidden layers.
\rt{} shares this spirit but differs structurally: rather than adding to activations at a manually chosen layer, we inject a synthesized vector as a full token position between the thinking delimiters, letting the complete transformer stack process it naturally.
Additionally, our focus is on the geometric structure specific to reasoning traces, and we intervene through orthogonal rotation rather than linear probing.
Supporting this design, \citet{esakkiraja2026therefore} show that LLMs encode action decisions in pre-generation activations with reasoning text serving as post-hoc rationalization, directly motivating the use of the input embedding as the seed for rotation into thinking space.

%% file: sections/3_preliminaries.tex
\section{Preliminaries}
\label{sec:prelims}


\begin{figure*} 
\centering
\includegraphics[width=\linewidth]{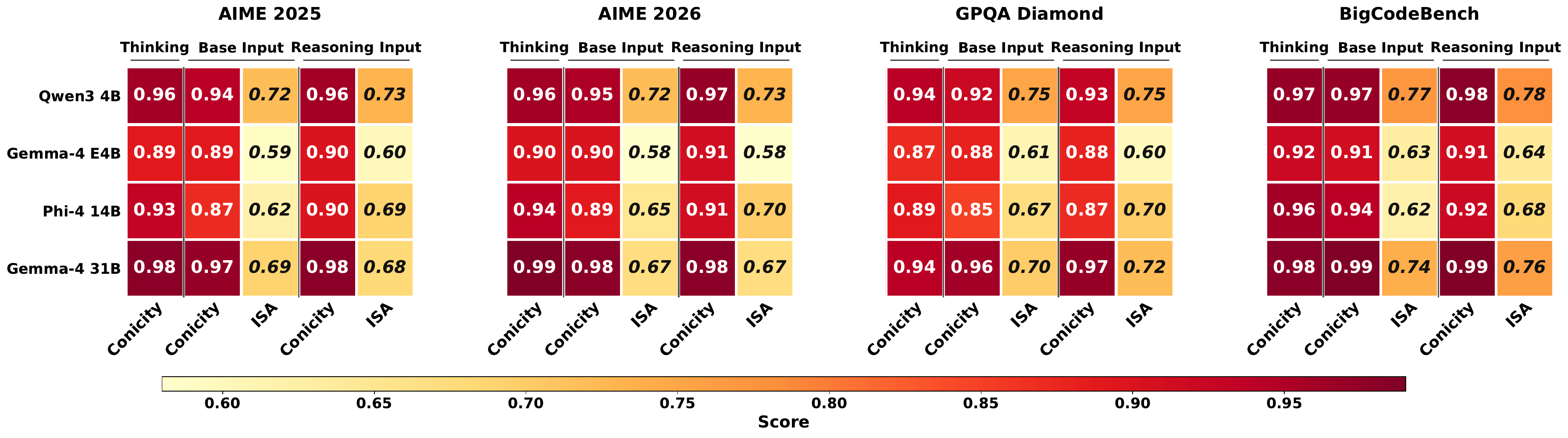}
\caption{Within-space conicity is high (avg. $\approx 0.93$), indicating tight cones in both thinking and input spaces. Cross-space ISA is markedly lower (avg. $\approx 0.66$), confirming that input and thinking cones are narrow but don't share an axis, the structure \rt{} exploits.}
\label{fig:conicity_results}
\end{figure*}


\subsection{Notation and Embedding Extraction}
\label{sec:notation}

Let $\mathcal{M}$ be a decoder-only transformer language model with $L$ layers and hidden dimension $D$.
For an input token sequence $\mathbf{x} = (x_1, \ldots, x_T)$, we write $\mathbf{h}^{(L)}(\mathbf{x}) \in \mathbb{R}^{T \times D}$ for the matrix of final-layer hidden states, where $\mathbf{h}^{(L)}_t(\mathbf{x})$ is the hidden state at position $t$.

\paragraph{Input embeddings.}
Given a query $q$ formatted as a chat-template prompt of $T$ tokens, the \emph{input embedding} is the mean of the final-layer hidden states:
\begin{equation}
    \ein(q) = \frac{1}{T} \sum_{t=1}^{T} \mathbf{h}^{(L)}_t(q) \;\in\; \mathbb{R}^D.
    \label{eq:input_emb}
\end{equation}
Computing $\ein(q)$ requires only a single forward pass through $\mathcal{M}$; no tokens are generated.

\paragraph{Thinking embeddings.}
Reasoning models generate an explicit reasoning trace delimited by special tokens like \texttt{<think>} and \texttt{</think>} \citep{yang2025qwen3, guo2025deepseek}.
Given query $q$, let $\mathbf{r} = (r_1, \ldots, r_S)$ denote the $S$ tokens generated within the thinking span.
The \emph{thinking embedding} is the mean of the final-layer hidden states at the thinking positions:
\begin{equation}
    \ethink(q) = \frac{1}{S} \sum_{s=1}^{S} \mathbf{h}^{(L)}_{|q|+s}\!\left([\mathbf{q};\, r_1, \ldots, r_S]\right) \;\in\; \mathbb{R}^D,
    \label{eq:think_emb}
\end{equation}
where $[\mathbf{q};\, r_1, \ldots, r_S]$ is the concatenated prompt-plus-thinking sequence and $|q|$ is the number of prompt tokens.

Computing $\ethink(q)$ uses a two-phase procedure:
(1) \textbf{Generation phase.} Run thinking-mode generation to produce the full response; record the token indices of the \texttt{<think>/</think>} boundaries in the generated sequence.
(2) \textbf{Extraction phase.} Run a single forward pass on the concatenated prompt-plus-thinking sequence to obtain $\mathbf{h}^{(L)}$, then mean-pool over the thinking-span positions.

\subsection{Conicity of Embedding Spaces}
\label{sec:conicity}

\citet{chandrahas2018geometry} introduced the notions of \emph{alignment to mean} (ATM) and \emph{conicity} to analyse the geometry of knowledge graph embedding spaces.
The \emph{alignment to mean} of a single vector $v$ with respect to a set $\mathcal{V}$ is its cosine similarity to the set mean, $\mathrm{ATM}(v, \mathcal{V}) = \cos(v, \boldsymbol{\mu}_\mathcal{V})$.
\emph{Conicity} is the average ATM over the set: given $\mathcal{V} = \{v_i\}_{i=1}^N \subset \mathbb{R}^D$ with mean $\boldsymbol{\mu}_\mathcal{V} = \tfrac{1}{N}\sum_i v_i$, the conicity of $\mathcal{V}$ is
\begin{equation}
    \conicity(\mathcal{V})
    = \frac{1}{N}\sum_{i=1}^{N} \cos\!\left(v_i,\;\boldsymbol{\mu}_\mathcal{V}\right),
    \label{eq:conicity}
\end{equation}
where $\cos(u,v) = u^\top v / (\|u\|\|v\|)$.

$\conicity(\mathcal{V}) \in [-1, 1]$; a value near $1$ indicates that all vectors point approximately in the direction of the mean, the set occupies a narrow \emph{cone} around $\boldsymbol{\mu}_\mathcal{V}$ rather than being spread across the sphere.

We also introduce a complementary metric, Inter-Space Alignment (ISA), defined as the cosine similarity between the centroids of two vector sets. 
Formally, for sets $\mathcal{P}$ and $\mathcal{Q}$ we define $\mathrm{ISA}(\mathcal{P}, \mathcal{Q}) = \cos(\boldsymbol{\mu}_\mathcal{P}, \boldsymbol{\mu}_\mathcal{Q})$, which quantifies the directional agreement between their mean representations.
Low ISA values signify directional divergence between $\mathcal{P}$ and $\mathcal{Q}$, suggesting that the aggregate representations of the two sets occupy distinct or unrelated regions of the latent space.

\subsection{The Orthogonal Procrustes Problem}
\label{sec:procrustes}

Given paired matrices $\mathbf{A} = [a_1, \ldots, a_N]^\top \in \mathbb{R}^{N \times D}$ and $\mathbf{B} = [b_1, \ldots, b_N]^\top \in \mathbb{R}^{N \times D}$, the \emph{orthogonal Procrustes problem}~\citep{schonemann1966generalized} seeks the orthogonal matrix $\mathbf{W} \in \mathbb{R}^{D \times D}$ (satisfying $\mathbf{W}^\top \mathbf{W} = \mathbf{I}_D$) that minimises the Frobenius-norm reconstruction error:
\begin{equation}
    \mathbf{W}^{*} = \argmin_{\mathbf{W}:\; \mathbf{W}^\top \mathbf{W} = \mathbf{I}_D}
    \left\|\mathbf{B} - \mathbf{A}\mathbf{W}^\top\right\|_F^2.
    \label{eq:procrustes}
\end{equation}
\citet{schonemann1966generalized} showed that the closed-form solution is obtained via the SVD of the cross-covariance matrix $\mathbf{M} = \mathbf{B}^\top \mathbf{A}$:
\begin{equation}
    \mathbf{M} = \mathbf{U}\,\boldsymbol{\Sigma}\,\mathbf{V}^\top
    \;\implies\;
    \mathbf{W}^{*} = \mathbf{U}\mathbf{V}^\top.
    \label{eq:procrustes_svd}
\end{equation}
The solution $\mathbf{W}^{*}$ is an orthogonal matrix.


\begin{figure}
     \centering
    \includegraphics[width=\linewidth]{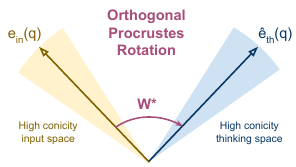}
    \caption{Schematic of the Rotate2Think method. An orthogonal Procrustes rotation $\mathbf{W}^{*}$ maps the mean-pooled input embedding $\ein(q)$ into thinking space. The resulting synthetic thinking vector $\ethr(q)$ is injected between thinking delimiters.}
    \label{fig:W_Star}
\end{figure}


\paragraph{Centering.}
In practice, we mean-center both matrices before fitting.
Let $\boldsymbol{\mu}_A$ and $\boldsymbol{\mu}_B$ be the column means of $\mathbf{A}$ and $\mathbf{B}$.
We apply Equation~\ref{eq:procrustes_svd} to the centered matrices $(\mathbf{A} - \mathbf{1}\boldsymbol{\mu}_A^\top)$ and $(\mathbf{B} - \mathbf{1}\boldsymbol{\mu}_B^\top)$, and account for the centroid shift at inference time:
\begin{equation}
    \ethr(q) = \mathbf{W}^{*}\!\left(\ein(q) - \boldsymbol{\mu}_A\right) + \boldsymbol{\mu}_B.
    \label{eq:transform}
\end{equation}

%% file: sections/4_method.tex
\section{\rt}
\label{sec:method}

\rt{} is a training-free, inference-time method that steers a language model toward its reasoning mode, with base models exhibiting ''thinking'' patterns without RL-based post-training and post-trained reasoning models showing improved reasoning.
At the heart of the approach is the geometric observation from \refsec{sec:conicity} and \reffig{fig:conicity_results} that both the input embedding space and the thinking embedding space are effectively one-dimensional cones, with distinct but consistently oriented mean directions.
This structure makes the mapping from one space to the other computable with very few examples and expressible as a single orthogonal rotation.

The method operates in two stages:
\begin{enumerate}
    \item \textbf{Fitting stage (offline).} Extract paired input and thinking embeddings for a set of training problems; fit an orthogonal Procrustes rotation $\mathbf{W}^*$ on the correctly-answered subset.
    \item \textbf{Inference stage (online).} For each new query, compute a synthetic thinking vector via the fitted rotation, inject it between thinking delimiters using \texttt{inputs\_embeds}, and generate the answer.
\end{enumerate}
\reffig{fig:W_Star} shows the schematic of \rt, while Algorithm~\ref{alg:rotate2think} gives the full procedure, and the following subsections describe each component.

\subsection{Rotation Fitting}
\label{sec:fitting}


\paragraph{Training example selection.}
We fit the Procrustes rotation using only problems the model answered correctly in thinking mode, on the hypothesis that correct solutions yield thinking vectors representative of productive reasoning, while incorrect ones may introduce noise that corrupts the rotation estimate.
When fitting from $K$ datasets other than the one being tested, $\mathcal{D}_1, \ldots, \mathcal{D}_K$, we sample up to 30 problems from each dataset and retain only those answered correctly in thinking mode.
Let $\mathcal{C}_k \subseteq \mathcal{D}_k$ denote this correct subset for dataset $k$.
We directly stack the individual input and thinking embeddings across all datasets into the paired matrices:
\begin{equation}
    \mathbf{A} = \bigl[e_{\mathrm{in}}(q_i)\bigr]_{\substack{k=1,\ldots,K \\ i \in \mathcal{C}_k}}^{\top},
    \mathbf{B} = \bigl[e_{\mathrm{th}}(q_i)\bigr]_{\substack{k=1,\ldots,K \\ i \in \mathcal{C}_k}}^{\top},
    \label{eq:ab_matrices}
\end{equation}
where $\mathbf{A}, \mathbf{B} \in \mathbb{R}^{N \times D}$ and $N = \sum_{k=1}^{K} |\mathcal{C}_k|$ is the total number of correctly answered examples across all source datasets.
Each row of $\mathbf{A}$ and the corresponding row of $\mathbf{B}$ form a paired $(e_{\mathrm{in}}, e_{\mathrm{th}})$ example used to fit the Procrustes rotation.


\paragraph{SVD fit.}
We apply the centered Procrustes solution (Equations~\ref{eq:procrustes}--\ref{eq:procrustes_svd}) to $\mathbf{A}$ and $\mathbf{B}$, obtaining $\mathbf{W}^*$, $\boldsymbol{\mu}_A$, and $\boldsymbol{\mu}_B$.

\subsection{Rotated Inference}
\label{sec:inference}

\paragraph{Synthetic thinking vector.}
Given a new test query $q$, we first compute $\ein(q)$ via a forward pass, then apply the rotation fitted in Equation~\ref{eq:transform} on $K$ given datasets other than the one being evaluated to obtain the synthetic thinking vector $\ethr(q)$.
This is a single matrix-vector multiply and incurs negligible overhead compared to a model forward pass.

\paragraph{Context construction and answer generation.}
We construct the \texttt{inputs\_embeds} tensor ($\mathbf{E}_{\mathrm{ctx}}$) for the generation call as a concatenation of four parts - $\mathbf{E}(q_{prompt})$, $\mathbf{E}(\texttt{<think>\textbackslash n})$, $\ethr(q)$, and $\mathbf{E}(\texttt{\textbackslash n</think>})$, 
where $\mathbf{E}(\cdot)$ is the model's token embedding lookup and $\ethr(q) \in \mathbb{R}^{1 \times D}$ occupies a single token position.
The thinking delimiters surrounding the injected vector ensure that the positional context matches the format the model encounters during thinking-mode generation.
We generate the answer by calling \texttt{model.generate(inputs\_embeds=$\mathbf{E}_{\mathrm{ctx}}$)}.

\begin{algorithm}[t]
\caption{\rt: fitting and inference}
\label{alg:rotate2think}
\begin{algorithmic}[1]
\Require Model $\mathcal{M}$, query $q$, pre-extracted paired embeddings $\{(\ein(q_i), \ethink(q_i))\}_{i=1}^N$ with correctness labels $\{y_i \in \{0,1\}\}$ (up to 30 examples per dataset) for rotation dataset list $\mathcal{D}_1,\ldots,\mathcal{D}_K$ with $\mathcal{C}_k$ being the list of correct examples from dataset $\mathcal{D}_k$ for $k=1,\ldots,K$
\Ensure Rotation parameters $(\mathbf{W}^*, \boldsymbol{\mu}_A, \boldsymbol{\mu}_B)$; answer for test query $q$
\Statex \textbf{-- Offline: Rotation Fitting --}
\State $\mathbf{A} \gets \bigl[e_{\mathrm{in}}(q_i)\bigr]_{\substack{k=1,\ldots,K \\ i \in \mathcal{C}_k}}^{\top}$ \hfill \Comment{Eq.~\ref{eq:ab_matrices}} \\
       $\mathbf{B} \gets \bigl[e_{\mathrm{th}}(q_i)\bigr]_{\substack{k=1,\ldots,K \\ i \in \mathcal{C}_k}}^{\top}$
       \hfill \Comment{Eq.~\ref{eq:ab_matrices}}
\State $\boldsymbol{\mu}_A \gets \mathrm{mean}(\mathbf{A})$;\quad $\boldsymbol{\mu}_B \gets \mathrm{mean}(\mathbf{B})$
\State $\mathbf{M} \gets (\mathbf{B} - \mathbf{1}\boldsymbol{\mu}_B^\top)^\top (\mathbf{A} - \mathbf{1}\boldsymbol{\mu}_A^\top)$ \\
       $\mathbf{U}\boldsymbol{\Sigma}\mathbf{V}^\top \gets \mathrm{SVD}(\mathbf{M})$
\State $\mathbf{W}^* \gets \mathbf{U}\mathbf{V}^\top$ \hfill \Comment{Eq.~\ref{eq:procrustes_svd}}
\Statex \textbf{-- Online: Inference for test query $q$ --}
\State $\ein(q) \gets$ forward pass on prompt tokens \Comment{no generation needed}
\State $\ethr(q) \gets \mathbf{W}^*(\ein(q) - \boldsymbol{\mu}_A) + \boldsymbol{\mu}_B$ \hfill \Comment{Eq.~\ref{eq:transform}}
\State $\mathbf{E}_{\mathrm{ctx}} \gets [\mathbf{E}(q_{\mathrm{prompt}}) \mid \mathbf{E}(\texttt{<think>\textbackslash n}) \mid \ethr(q) \mid \mathbf{E}(\texttt{\textbackslash n</think>})]$
\State \Return $\mathcal{M}.\texttt{generate}(\mathbf{E}_{\mathrm{ctx}})$
\end{algorithmic}
\end{algorithm}

%% file: sections/5_experiments.tex
\section{Experimental Setup}
\label{sec:experiments}


\subsection{Datasets}
\label{sec:data}

We evaluate on five benchmarks spanning mathematics, science, code, and visual-mathematics chosen to represent diverse reasoning domains.

\paragraph{AIME 2025 and AIME 2026.}
The American Invitational Mathematics Examination (AIME) is a competition-level mathematics benchmark consisting of 30 integer-answer problems per year, drawn from number theory, algebra, combinatorics, and geometry.
Problems require multi-step reasoning and yield an integer answer in $[0, 999]$.
We use the official 2025 and 2026 sets, treating each year as a separate benchmark \citep{aime}.

\paragraph{GPQA Diamond.}
GPQA \citep{rein2023gpqa} is a multiple-choice benchmark of graduate-level science questions authored by domain experts.
We evaluate on the Diamond split (198 questions spanning Biology, Chemistry, and Physics), the hardest subset, which requires deep subject-matter knowledge and eliminates most surface-level shortcuts.

\paragraph{BigCodeBench.}
BigCodeBench \citep{zhuo2024bigcodebench} is a function-level Python code generation benchmark with 1,140 diverse tasks covering library usage, data manipulation, and algorithm implementation.
Each task is evaluated by executing the generated code against a suite of unit tests.

\paragraph{\mv.}
MATH-Vision \citep{mathvision} is a multimodal mathematical problem-solving benchmark with 3040 visual reasoning problems in the full test set. We use the \texttt{testmini} split of the dataset containing 304 problems.


\subsection{Baselines and Evaluation Protocol}
\label{sec:eval}

We report two metrics for each condition: \textbf{Acc} (accuracy, \%), and \textbf{Tok} (average number of tokens generated per problem).
For each model-benchmark pair, we evaluate four modes:

\begin{itemize}
    \item \textbf{Base.} Standard (non-thinking) generation: the model generates its answer directly without producing a reasoning trace.

    \item \textbf{Reasoning.} Full thinking-mode generation: the model produces a complete reasoning trace before its answer.

    \item \textbf{Base+R2T.} The rotation is fitted from the base model variant's input and to the reasoning model's thinking embeddings. During inference, the base model receives the synthetic thinking vector injected between thinking delimiters.

    \item \textbf{Reasoning+R2T.} The rotation is fitted from the reasoning model's embeddings. The same reasoning model is used for inference, with the synthetic thinking vector injected between thinking delimiters.
\end{itemize}


\begin{table*}[t]
\centering
\footnotesize

\begin{tabular}{l|l||cc|cc||cc|cc}
\toprule
& & \multicolumn{2}{c}{\textbf{Base}} & \multicolumn{2}{c}{\textbf{Base+R2T}} & \multicolumn{2}{c}{\textbf{Reasoning}} & \multicolumn{2}{c}{\textbf{Reasoning+R2T}} \\
\cmidrule(lr){3-4}\cmidrule(lr){5-6}\cmidrule(lr){7-8}\cmidrule(lr){9-10}
\textbf{Benchmark} & \textbf{Model} & Acc \% & Tok & Acc \% & Tok & Acc \% & Tok & Acc \% & Tok \\
\midrule
\multirow{4}{*}{AIME 2025}
  & Qwen3 4B           & 43.33 & \textit{10444} & \textbf{46.67} & \textit{8299} & 70.00 & \textit{20944} & \textbf{80.00} & \textit{20894} \\
  & Gemma-4 E4B        & 36.67 & \textit{3438} & \textbf{40.00} & \textit{5408} & 40.00 & \textit{6968} & \textbf{50.00} & \textit{5891} \\
  & Phi-4 14B        & 16.67 & \textit{1672} & \textbf{20.00} & \textit{1639} & 73.33 & \textit{15053} & \textbf{76.67} & \textit{17108} \\
  & Gemma-4 31B        & 70.00 & \textit{2927} & \textbf{73.33} & \textit{4110} & 83.33 & \textit{7170} & \textbf{86.67} & \textit{4240} \\
\midrule
\multirow{4}{*}{AIME 2026}
  & Qwen3 4B           & 40.00 & \textit{9359} & \textbf{53.33} & \textit{8855} & 70.00 & \textit{20609} & \textbf{73.33} & \textit{21076} \\
  & Gemma-4 E4B        & 33.33 & \textit{3691} & \textbf{46.67} & \textit{5422} & 43.33 & \textit{6829} & \textbf{53.33} & \textit{5586} \\
  & Phi-4 14B        & 10.00 & \textit{1380} & \textbf{16.67} & \textit{1515} & 76.67 & \textit{17077} & \textbf{80.00} & \textit{17540} \\
  & Gemma-4 31B        & 80.00 & \textit{2119} & \textbf{83.33} & \textit{2036} & 83.33 & \textit{7894} & \textbf{86.67} & \textit{3098} \\
\midrule
\multirow{4}{*}{GPQA Diamond}
  & Qwen3 4B           & 46.67 & \textit{737} & \textbf{48.48} & \textit{549} & 67.17 & \textit{8018} & \textbf{69.19} & \textit{9126} \\
  & Gemma-4 E4B        & 55.07 & \textit{1979} & \textbf{64.00} & \textit{3254} & 58.67 & \textit{3646} & 60.33 & \textit{3342} \\
  & Phi-4 14B        & 54.55 & \textit{658} & \textbf{56.57} & \textit{746} & 69.19 & \textit{11353} & \textbf{69.70} & \textit{11569} \\
  & Gemma-4 31B        & 74.24 & \textit{1284} & \textbf{78.28} & \textit{1390} & 84.34 & \textit{5728} & \textbf{86.87} & \textit{5614} \\
\midrule
\multirow{4}{*}{BigCodeBench}
  & Qwen3 4B           & 38.24 & \textit{409} & \textbf{38.51} & \textit{404} & 41.58 & \textit{5787} & \textbf{41.84} & \textit{5670} \\
  & Gemma-4 E4B        & 36.67 & \textit{608} & \textbf{38.38} & \textit{1238} & 36.84 & \textit{1227} & \textbf{37.98} & \textit{1252} \\
  & Phi-4 14B        & \textbf{39.91} & \textit{242} & 39.74 & \textit{582} & 43.68 & \textit{4172} & \textbf{44.03} & \textit{3962} \\
  & Gemma-4 31B        & 46.49 & \textit{364} & \textbf{47.54} & \textit{1577} & \textbf{48.51} & \textit{2198} & 48.25 & \textit{2291} \\

\bottomrule
\end{tabular}
\caption{Accuracy (\%) and average tokens generated (Tok). \rt{} improves accuracy in 30 of 32 model–benchmark configurations, with comparable token budgets. R2T = \rt.}
\label{tab:main_results}
\setlength{\tabcolsep}{4pt}
\end{table*}



\subsection{Models Evaluated}
\label{sec:models}

We evaluate four language models spanning three families and a range of parameter counts.

\paragraph{Qwen3 4B.}
\texttt{Qwen/Qwen3-4B-Instruct-2507} (which we use as base model) and \texttt{Qwen/Qwen3-4B-Thinking-2507} (which we use as reasoning model) \citep{yang2025qwen3} are 4B-parameter dense language models.

\paragraph{Phi-4 14B.}
We use \texttt{microsoft/phi-4} as the base model and \texttt{microsoft/Phi-4-reasoning-plus} as the reasoning model. These are 14B, dense models \citep{abdin2024phi4}.

\paragraph{Gemma-4 E4B and 31B.}
We use \texttt{google/gemma-4-E4B-it} (8B MoE model) and \texttt{google/gemma-4-31B-it} (31B dense model) \citep{google2025gemma4} in our evaluation.

The reasoning versions of Qwen3 4B, and Phi-4 14B use \texttt{<think>}/\texttt{</think>} delimiters to bracket the reasoning trace.
Gemma-4 models, on the other hand, use the \texttt{<|think|>} flag to toggle between non-thinking (which we use as base model for Gemma-4) and thinking (which we use as reasoning model for Gemma-4) modes with model-specific thinking delimiters (\texttt{<|channel>}/\texttt{<channel|>}) and enforce an exclusive-or constraint between \texttt{input\_ids} and \texttt{inputs\_embeds}. We handle this via a forward hook that replaces a designated placeholder token's embedding with the synthetic thinking vector during prefill.
We use the \texttt{4-bit} quantized version of Gemma-4 31B model.

%% file: sections/6_results.tex
\section{Results}
\label{sec:results}


\noindent \textbf{Geometric structure of reasoning hidden states.} \reffig{fig:conicity_results} reports conicity and ISA scores across all models and benchmarks.
Conicity is uniformly high across all conditions (0.87–0.99), confirming that reasoning model hidden states occupy a narrow cone in the representation space, which is a prerequisite for the low-rank geometric characterization that Rotate2Think exploits.
ISA values between input and thinking embeddings are substantially lower on average (0.659 base, 0.670 reasoning), 
indicating that input and thinking cones are individually narrow but do not share an axis, motivating our Procrustes-based rotation.
Both properties hold consistently across benchmarks and model families, suggesting that high conicity and low ISA are stable geometric properties of current reasoning models.

\noindent \textbf{Base $\to$ Base+\rt.} 
Injecting a synthetic thinking vector into base models (Base+R2T) yields consistent accuracy gains across every model and benchmark tested, as seen in \reftbl{tab:main_results}. AIME improvements reach up to $+13$ percentage points. On GPQA Diamond and AIME 2026, Gemma-4 E4B under Base+R2T surpasses the same model in full reasoning mode. These results suggest the geometric vector carries a genuine reasoning signal independent of the chain-of-thought length.
Beyond accuracy, we qualitatively observe that Qwen3-4B under \rt{} produces self-correcting phrases such as ''Wait\ldots'' and ''Let me reconsider\ldots'' — patterns characteristic of RL-trained reasoning models but absent in standard base-model generation. This behavior does not appear consistently across Phi-4 and Gemma-4, so we report it as an incidental observation rather than a general finding.

\noindent \textbf{Reasoning $\to$ Reasoning+\rt.} 
As seen in \reftbl{tab:main_results}, augmenting reasoning models with the \rt{} primer yields consistent accuracy improvements across all models and benchmarks, with gains present regardless of model scale or task domain. 
On AIME, improvements range from modest to substantial with Gemma-4~E4B gaining $+10$ percentage points on both AIME~2025 and AIME~2026, while Qwen3~4B improves by $+10$ and $+3.3$ points, respectively. GPQA Diamond and BigCodeBench follow the same trend, with almost all models registering gains under Reasoning+R2T. Notably, these improvements occur without a commensurate increase in token generation, suggesting that R2T does not prompt the model to reason more, but rather more effectively. The synthetic thinking vector appears to prime the model's internal reasoning process, improving the quality of the chain-of-thought that follows.


\begin{figure} 
\centering
\includegraphics[width=\linewidth]{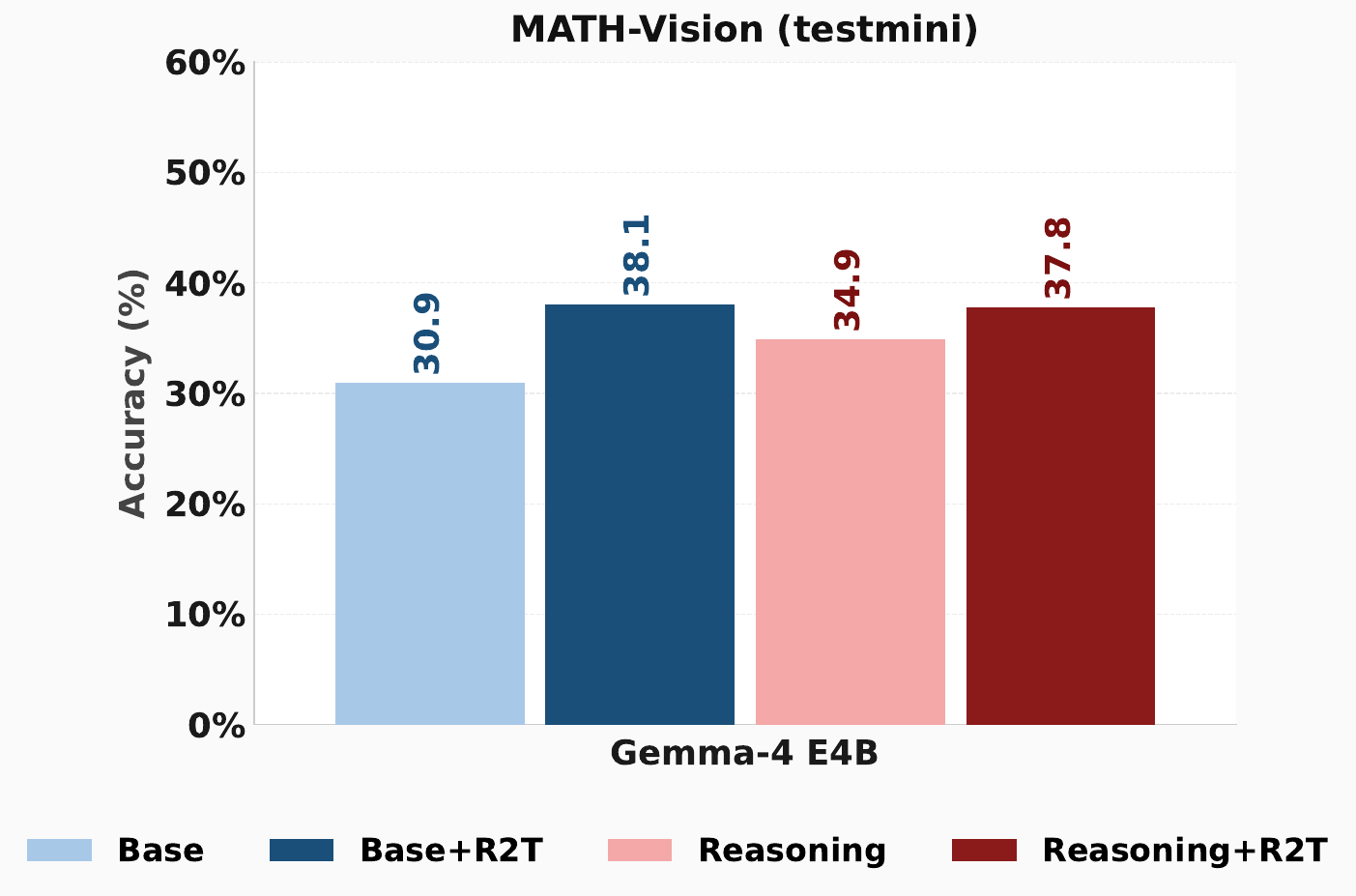}
\caption{\rt{} results on \mv{} for Gemma-4 E4B. The rotation is fit on text benchmarks only and applied zero-shot to a multimodal benchmark. Base+R2T ($38.1$) exceeds Reasoning ($34.9$) and approaches Reasoning+R2T ($37.8$), described further in \refsec{sec:results}.}
\label{fig:r2t_results_vision}
\end{figure}


\noindent \textbf{Generalizing to Multimodal Reasoning.} 
To examine whether \rt{} transfers beyond language-only inputs, we evaluate on \mv{} using Gemma-4 E4B, with the rotation matrix fit on 30 examples each from the four text benchmarks (AIME 2025, AIME 2026, GPQA Diamond, BigCodeBench) — without any multimodal data or modality-specific adaptation.
As shown in \reffig{fig:r2t_results_vision}, \rt{} improves accuracy for both the base model and the reasoning model, consistent with the pattern observed on text benchmarks. That these gains carry over to a multimodal LLM (MLLM) without any modification suggests \rt{} may extend naturally to multimodal settings, broadening its practical applicability beyond language-only reasoning.


\section{Discussion}
\label{sec:discussion}

Taken together, the results in \refsec{sec:results} support a coherent picture of what \rt{} is doing.

\noindent \textbf{The thinking embedding is a meaningful priming signal.}
Across diverse model families and parameter scales, rotating the input embedding into thinking space and injecting a single token consistently steers the model toward stronger reasoning behavior on hard problems.
This is consistent with the geometric hypothesis in \refsec{sec:conicity}: if both input and thinking embeddings form tight cones with non-aligned axes, then a single rotation can reliably shift the model's context from one regime to the other.

\noindent \textbf{Reasoning capacity may be latent and geometrically accessible.}
\rt{} intervenes at a single point (one rotated embedding, one injected token) with no gradient updates, no additional training, and no modification to model weights.
That such a minimal perturbation produces consistent accuracy gains across both base and reasoning models hints that the injected thinking vector may provide a useful initialization of the reasoning process, steering the model toward a representational regime where reasoning is more reliably expressed.

\noindent \textbf{Thinking-space geometry transcends input modality.} 
The transfer of \rt{} to \mv, with the rotation fit purely on text reasoning, suggests that the thinking-space geometry is a property of the model rather than of the input modality. That such a transfer occurs without any multimodal data or modality-specific refitting is consistent with the Platonic Representation Hypothesis \citep{huh2024platonic}, and hints that the learned rotation may capture a modality-agnostic signature of the model's reasoning regime.

%% file: sections/7_conclusion.tex
\section{Conclusion}
\label{sec:conclusion}

We introduced \rt{}, a training-free method that improves language model reasoning by rotating the mean-pooled input embedding into “thinking space” via orthogonal Procrustes analysis and injecting the result as a single token between thinking delimiters.
Our key geometric insight is that input and thinking embeddings both exhibit high conicity, i.e., they occupy narrow cones in representation space, but with systematically misaligned mean directions, reducing the cross-space mapping to a pure rotation.
\rt{} consistently improves accuracy for both base and reasoning models across benchmarks and LLMs evaluated. 
Notably, \rt{} also extends to multimodal settings as evaluated on \mv, it improves reasoning accuracy in MLLMs without any modality-specific adaptation, suggesting that the thinking-space geometry exploited by \rt{} may not be exclusive to language-only pretraining, and could reflect a more general structural property of the representation space.

%% file: sections/8_limitations.tex
\section*{Limitations}

There are limitations of the current work that warrant acknowledgment.
First, \rt{} in its current form requires access to a reasoning-capable variant of the target model to extract thinking embeddings; it cannot be applied to model families for which no such variant exists.
Second, while gains are consistent across mathematical reasoning and scientific knowledge benchmarks, improvements on code generation tasks are modest, suggesting the method's effectiveness is tied to task domains where extended deliberation is most consequential.
Third, all evaluations are conducted under a single-token injection scheme; whether more expressive interventions, such as injecting a sequence of rotated embeddings, for instance, could yield further gains is an open question.
Finally, \rt{} operates directly on hidden state representations and therefore requires white-box access to the model's internals, specifically, the last hidden state of the input sequence. This precludes application to closed-source models, where only token-level outputs are accessible, and limits the method to open-weight models for which activation extraction is feasible.
We hope this perspective encourages further work at the intersection of representational geometry and inference-time reasoning, and that the lightweight nature of \rt{} makes it a practical building block for future reasoning systems.

%% file: sections/10_appendix.tex
\appendix

\section{Appendix: Hyperparameters}
\label{sec:hyperparameters}

\subsection{Generation}

All single-run experiments, except Phi-4 (base), use sampling. Sampling parameters follow each model family's recommended settings from their respective \texttt{huggingface} model cards and are listed in Table~\ref{tab:sampling_params}.

\begin{table*}[h]
\centering
\begin{tabular}{lccc}
\toprule
\textbf{Model} & \textbf{temperature} & \textbf{top\_k} & \textbf{top\_p} \\
\midrule
Gemma-4 E4B / 31B & 1.0 & 64 & 0.95 \\
Qwen3 4B (base)   & 0.7 & 20 & 0.80 \\
Qwen3 4B (reasoning) & 1.0 & 20 & 0.95 \\
Phi-4 14B (base) & -- & -- & -- \\
Phi-4 14B (reasoning) & 0.6 & 20 & 0.95 \\
\bottomrule
\end{tabular}
\caption{Sampling hyperparameters used per model family.}
\label{tab:sampling_params}
\end{table*}

\subsection{Rotation Fitting}

The orthogonal Procrustes rotation $\mathbf{W}^*$ is fitted offline using a deterministic SVD (no gradient descent, no learning rate). Fitting proceeds as follows:
\begin{itemize}
    \item At most \textbf{30} examples per dataset are considered (\texttt{MAX\_ROTATION\_SAMPLES = 30}).
    \item Only correctly-answered examples are retained.
    \item The mean input embedding and mean thinking embedding are computed per dataset over the retained examples.
    \item These per-dataset means are stacked and used to fit a single Procrustes rotation with centroid alignment.
\end{itemize}

For the multimodal transfer experiment (MATH-Vision), the rotation is fitted on \textbf{120 examples} in total (4 text datasets $\times$ 30 samples each: AIME 2025, AIME 2026, GPQA Diamond, and BigCodeBench), and applied zero-shot to visual problems.

\subsection{Quantization}

Gemma-4 31B is loaded in \textbf{4-bit} (NF4 + double quantization, \texttt{bfloat16} compute dtype) via \texttt{bitsandbytes}. All other models are loaded in \texttt{bfloat16} without quantization.

\subsection{Embedding Extraction}

Embeddings are obtained by mean-pooling the last-layer hidden states over all non-padding token positions. Procrustes fitting is performed in \texttt{float64}; injected synthetic thinking vectors are cast to \texttt{bfloat16} for inference.